\documentclass[11pt,a4paper]{article}
\usepackage{varioref}
\usepackage[hyperref]{acl2020}
\usepackage{times}
\usepackage{latexsym}

\usepackage{microtype}
\usepackage{graphicx}
\usepackage{multirow}
\usepackage{soul}
\usepackage{amssymb}
\usepackage{amsmath}
\usepackage{amsthm}
\usepackage{cleveref}
\crefformat{section}{\S#2#1#3}
\crefformat{subsection}{\S#2#1#3}
\crefformat{subsubsection}{\S#2#1#3}

\aclfinalcopy 
\def\aclpaperid{1966} 

\title{Clinical Reading Comprehension: \\A Thorough Analysis of the emrQA Dataset}

\author{Xiang Yue \\
  \And
  Bernal Jimenez Gutierrez \\
  The Ohio State University \\
  \texttt{\{yue.149, jimenezgutierrez.1, sun.397\}@osu.edu} \\\And
  Huan Sun
}

\date{}

\begin{document}
\maketitle

\begin{abstract}
Machine reading comprehension has made great progress in recent years owing to large-scale annotated datasets. In the clinical domain, however, creating such datasets is quite difficult due to the domain expertise required for annotation. Recently, \citet{pampari2018emrqa} tackled this issue by using expert-annotated question templates and existing i2b2 annotations to create \textit{emrQA}, the first large-scale dataset for question answering (QA) based on clinical notes. In this paper, we provide an in-depth analysis of this dataset and the clinical reading comprehension (CliniRC) task. From our qualitative analysis, we find that (i) emrQA answers are often incomplete, and (ii) emrQA questions are often answerable without using domain knowledge. From our quantitative experiments, surprising results include that (iii) using a small sampled subset (5\%-20\%), we can obtain roughly equal performance compared to the model trained on the entire dataset, (iv) this performance is close to human expert's performance, and (v) BERT models do not beat the best performing base model. Following our analysis of the emrQA, we further explore two desired aspects of CliniRC systems: the ability to utilize clinical domain knowledge and to generalize to unseen questions and contexts. We argue that both should be considered when creating future datasets.\footnote{Our code is available at \url{https://github.com/xiangyue9607/CliniRC}.} 
\end{abstract}

\section{Introduction}
Medical professionals often query over clinical notes in Electronic Medical Records (EMRs) to find information that can support their decision making \cite{demner2009can,rosenbloom2011data,wang2018clinical}. One way to facilitate such information seeking activities is to build a natural language question answering (QA) system that can extract precise answers from clinical notes \cite{cairns2011mipacq,cao2011askhermes,wren2011question,abacha2016recognizing,BenAbacha2019}. 

\begin{figure}[!t]
    \centering
    \includegraphics[width=\linewidth]{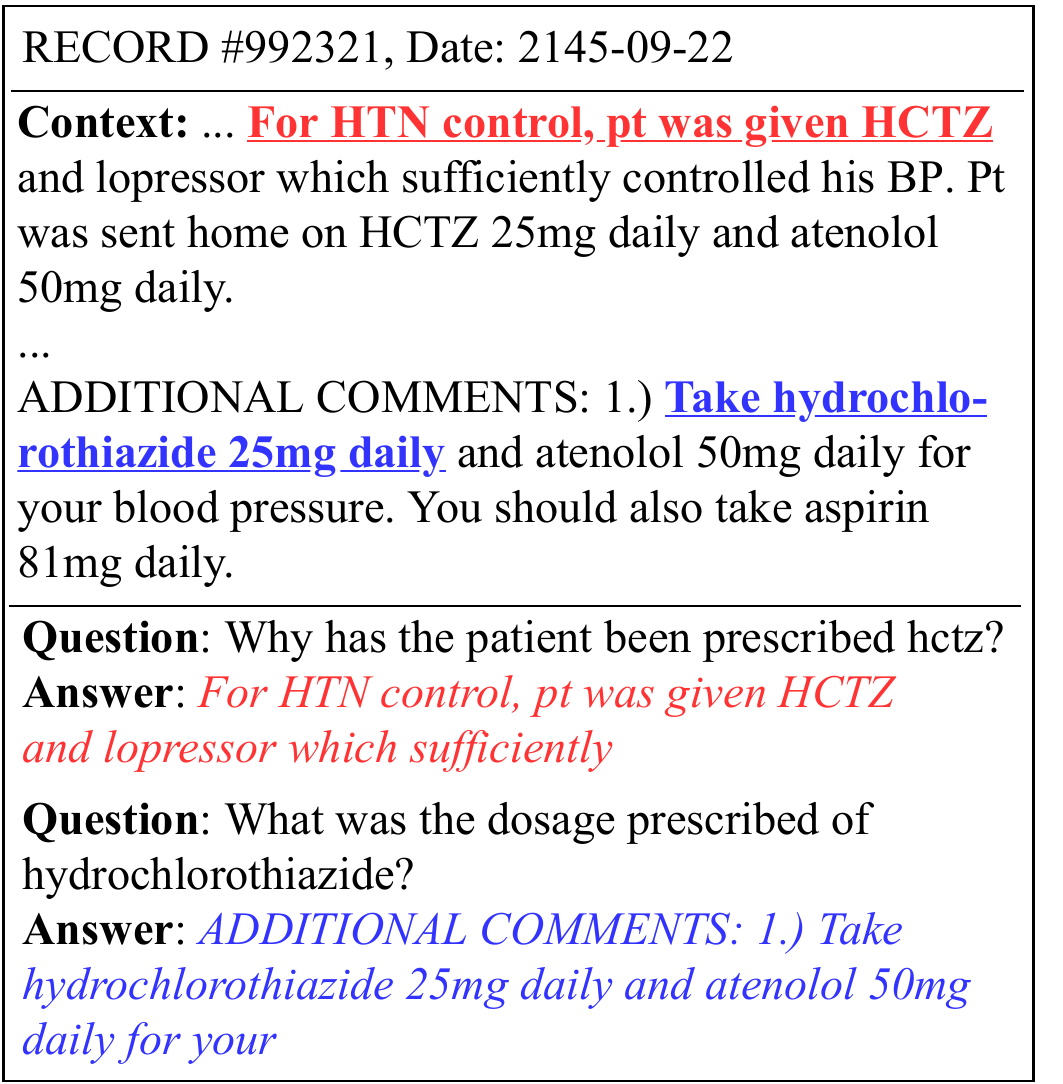}
    \caption{Examples from the emrQA dataset: Part of a clinical note as \textit{context} and 2 \textit{question-answer} pairs. Due to the original emrQA generation issues, oftentimes answers are incomplete or contain irrelevant parts to the questions (the underlined parts are what we think the most relevant to the questions).}
    \label{fig:intro_example}
\end{figure}

Machine reading comprehension (RC) aims to automatically answer questions based on a given document or text corpus and has drawn wide attention in recent years. Many neural models \cite{cheng2016long,wang2017gated,Wang2016MachineCU,Seo2017Bidirectional,chen2017reading,devlin2019bert} have achieved very promising results on this task, owing to large-scale QA datasets \cite{hermann2015teaching,rajpurkar2016squad,trischler-etal-2017-newsqa,joshi-etal-2017-triviaqa,yang2018hotpotqa}. Unfortunately, clinical reading comprehension (CliniRC) has not observed as much progress due to the lack of such QA datasets. 

In order to create QA pairs on clinical texts, annotators must have considerable medical expertise and data handling must be specifically designed to address ethical issues and privacy concerns. Due to these requirements, using crowdsourcing like in the open domain to create large-scale clinical QA datasets becomes highly impractical \cite{wei2018clinical}.

Recently, \citet{pampari2018emrqa} found a smart way to tackle this issue and created \textit{emrQA}, the first large-scale QA dataset on clinical texts. Instead of relying on crowdsourcing, emrQA was semi-automatically generated based on annotated question templates and existing annotations from the n2c2 (previously called i2b2) challenge datasets\footnote{https://portal.dbmi.hms.harvard.edu/projects/n2c2-nlp/}. Example QA pairs from the dataset are shown in Figure \ref{fig:intro_example}.

In this paper, we aim to gain a deep understanding of the CliniRC task and conduct a thorough analysis of the emrQA dataset. We first explore the dataset directly by carrying out a meticulous qualitative analysis on randomly-sampled QA pairs and we find that: 1) Many answers in the emrQA dataset are incomplete and hence are hard to read and ineffective for training (\cref{subsec:answer_quality}). 2) Many questions are simple: More than 96\% of the examples contain the same key phrases in both questions and answers. Though \citet{pampari2018emrqa} claims that 39\% of the questions may need knowledge to answer, our error analysis suggests only a very small portion of the errors (2\%) made by a state-of-the-art reader might be due to missing external domain knowledge (\cref{subsec:question_difficulty}).

Following our qualitative analysis of the emrQA dataset, we conduct a comprehensive quantitative analysis based on state-of-the-art readers and BERT models (BERT-base \cite{devlin2019bert} as well as its biomedical and clinical versions: BioBERT \cite{Lee2019biobert} and ClinicalBERT \cite{alsentzer2019publicly}) to understand how different systems behave on the emrQA dataset.  Surprising results include: 1) Using a small sampled subset (5\%-20\%), we can obtain roughly equal performance compared to the model trained on the entire dataset, suggesting that many examples in the dataset are redundant (\cref{subsec:sample_subset}). 2) The performance of the best base model is close to the human expert's performance\footnote{Which is obtained by comparing emrQA answers to answers created by our medical experts on sampled QA pairs.} (\cref{subsec:ceiling}). 3) The performance of BERT models is around 1\%-5\% worse than the best performing base model (\cref{subsec:worse_BERT}). 

After completing our analysis of the dataset, we explore two potential needs for systems doing CliniRC: 1) The need to represent and use clinical domain knowledge effectively (\cref{subsec:external_knowledge}) and 2) the need to generalize to unseen questions and contexts (\cref{subsec:generalizability}). To investigate the first one, we analyze several types of clinical questions that require domain knowledge and can frequently appear in the real clinical setting. We also carry out an experiment showing that adding knowledge explicitly yields around 5\% increase in F1 over 
the base model when tested on samples that we created by altering the original questions to involve semantic relations. To study generalizability, we ask medical experts to create new questions based on the unseen clinical notes from MIMIC-III \cite{mimiciii}, a freely accessible critical care database. We find that the performance of the best model trained on emrQA drops by 40\% under this new setting, showing how critical it is for us to develop more robust and generalizable models for the CliniRC task. 

In summary, given our analysis of the emrQA dataset and the task in general, we conclude that future work still needs to create better datasets to advance CliniRC. Such datasets should be not only large-scale, but also less noisy, more diverse, and allow researchers to directly evaluate a system's ability to encode domain knowledge and to generalize to new questions and contexts.

\begin{table}[!t]
\begin{tabular}{lrr}
\hline
 & \textbf{Medication} & \textbf{Relation} \\ \hline
\# Question & 222,957 & 904,592 \\ 
\# Context & 261 & 423 \\ \hline
\# Question Template & 80 & 139 \\ \hline
Question: avg. tokens & 8.00 & 7.91 \\
Answers: avg. tokens & 9.47 & 10.41 \\
Context: avg. tokens & 1062.66 & 889.23 \\ \hline
\end{tabular}
\caption{Statistics of two major subsets, \textit{Medication} and \textit{Relation}, of the emrQA dataset.}
\vspace{-10pt}
\label{tbl: statistics}
\end{table}

\section{Overview of the emrQA dataset}
Similar to the open-domain reading comprehension task, the Clinical Reading Comprehension (CliniRC) task is defined as follows: 
\theoremstyle{definition}
\newtheorem{definition}{Definition}[section]
\begin{definition}
Given a patient's clinical note (context) $C=\{c_1,..., c_n\}$ and a question $Q =\{t_1, ..., t_m\}$, the CliniRC task aims to extract a continuous span $A = \{c_i, c_{i+1},..., c_{i+k}\}(1 \leq i \leq i + k \leq n)$ from the context as the answer, where $c_i, t_j$ are tokens.
\end{definition}

\begin{figure}
    \centering
    \includegraphics[width=\linewidth]{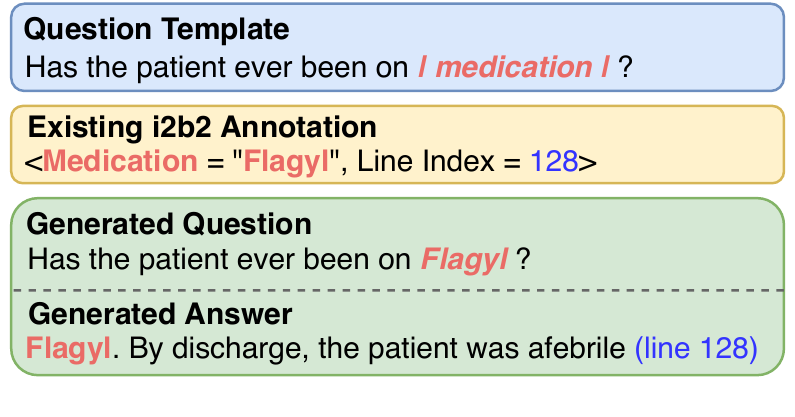}
    \caption{An example to illustrate how emrQA generates QA pairs.}
    \vspace{-10pt}
    \label{fig:emrQA_generation}
\end{figure}

The emrQA dataset \cite{pampari2018emrqa} was semi-automatically generated from expert-annotated question templates and existing i2b2 annotations. More specifically, clinical question templates were first created by human experts. Then, manual annotations from the medication information extraction, relation learning, and coreference resolution i2b2 challenges were re-framed into answers for the question templates. After linking question templates to i2b2 annotations, the gold annotation entities were used to both replace placeholders in the question templates and extract the sentence around them as answers. An example of this generation process can be seen in Figure \ref{fig:emrQA_generation}.

The emrQA dataset contains 5 subsets: \textit{Medication}, \textit{Relation}, \textit{Heart Disease}, \textit{Obesity} and \textit{Smoking}, which were generated from 5 i2b2 challenge datasets respectively. The answer format in each dataset is different. For the \textit{Obesity} and \textit{Smoking} datasets, answers are categorized into 7 classes and the task is to predict the question's class based on the context. For the \textit{Medication}, \textit{Relation}, and \textit{Heart Disease} datasets, answers are usually short snippets from the text accompanied by a longer span around it which we refer to as an evidence. The short snippet is a single entity or multiple entities while the evidence contains the entire line around those entities in the clinical note. For questions that cannot be answered via entities, only the evidence is provided as an answer. Given that some questions do not have short answers and that entire evidence spans are usually important for supporting clinical decision making \cite{demner2009can}, we treat the \textit{answer evidence} \footnote{For simplicity, we use ``answer" directly henceforth.} as our answer just as is done in \cite{pampari2018emrqa}.

In this work, we mainly focus on the \textit{Medication} and \textit{Relation} datasets because (1) they make up 80\% of the entire emrQA dataset and (2) their format is consistent with the span extraction task, which is more challenging and meaningful for clinical decision making support. We filter the answers whose lengths (number of tokens) are more than 20. The detailed statistics of the two datasets are shown in Table \ref{tbl: statistics}.

\section{In-depth Qualitative Analysis}
In this section, we carry out an in-depth analysis of the emrQA dataset. We aim to examine (1) the quality and (2) level of difficulty for the generated QA pairs in the emrQA dataset. 

\subsection{How clean are the emrQA answers?}
\label{subsec:answer_quality}
Since the emrQA dataset was created via a generation framework unlike human-labeled or crowdsourcing datasets, the quality of the datasets remains largely unknown. In order to use this dataset to explore the CliniRC task, it is essential to determine whether it is meaningful.

In order to do this, we randomly sample 50 QA pairs from the \textit{Medication} and the \textit{Relation} datasets respectively. Since some questions share the same answer due to automatic generation, we make sure all the samples have different answers.

\begin{table}[t]
\centering
\begin{tabular}{cccc}
\hline
 & Metric & \textbf{Medication} & \textbf{Relation} \\ \hline \hline
& Quality Score & 3.92 & 4.75 \\ 
 & EM & 26.0 & 92.0 \\
 & F1 & 74.7 & 95.4 \\ \hline
\end{tabular}
\caption{An estimate of the quality of answers in the \textit{Medication} and \textit{Relation} datasets based on the analysis of our randomly sampled 50 questions for each dataset. Quality scores are the average of two human annotators' (maximum: 5). EM and F1 scores are calculated between \textit{human-labeled} answers v.s. \textit{emrQA} answers.}
\label{tbl:dataset_quality}
\end{table}

\begin{table*}[t]
\resizebox{\textwidth}{!}{%
\begin{tabular}{llllcc}
\hline
\multirow{2}{*}{\textbf{Error Type}} & \multirow{2}{*}{\textbf{Question}} & \multirow{2}{*}{\textbf{emrQA Answers}} & \multirow{2}{*}{\textbf{Prediction}} & \multicolumn{2}{c}{\textbf{Error Ratio}} \\ \cline{5-6} 
 &  &  &  & \textbf{Medication} & \textbf{Relation} \\ \hline
\textbf{\begin{tabular}[c]{@{}l@{}}Span mismatch\\ - include key info\end{tabular}} & \begin{tabular}[c]{@{}l@{}}
Does she have a \\history of known\\ drug allergies?
\end{tabular} & \begin{tabular}[c]{@{}l@{}}
ALLERGIES: \\ He had no known \\drug allergies
\end{tabular} & \begin{tabular}[c]{@{}l@{}}
 He had no known \\drug allergies
\end{tabular} & \textbf{78\%} & \textbf{66\%} \\ \hline
\textbf{\begin{tabular}[c]{@{}l@{}}Span mismatch \\ - miss key info\end{tabular}} & \begin{tabular}[c]{@{}l@{}}What is the current \\ dose of lasix?\end{tabular} & \begin{tabular}[c]{@{}l@{}}MEDS: K-Dur 20 BID, \\ Nexium 20, lasix 160 BID\end{tabular} & BID & \textbf{4\%} & \textbf{0\%} \\ \hline
\textbf{\begin{tabular}[c]{@{}l@{}}Ambigious \\ questions\end{tabular}} & \begin{tabular}[c]{@{}l@{}}What is the patient's \\ low history?\end{tabular} & \begin{tabular}[c]{@{}l@{}}At the time of discharge, \\ her potassium had been \\ low despite repletion\end{tabular} & 11) Low grade,anemia & \textbf{8\%} & \textbf{4\%} \\ \hline
\textbf{\begin{tabular}[c]{@{}l@{}}Incorrect \\ golds\end{tabular}} & \begin{tabular}[c]{@{}l@{}}What is the patient's\\  incisions status?\end{tabular} & \begin{tabular}[c]{@{}l@{}}Wash incisions with warm \\ water and gentle soap\end{tabular} & \begin{tabular}[c]{@{}l@{}}Do not apply lotions, \\ creams, ointments or \\ powders to incision\end{tabular} & \textbf{2\%} & \textbf{2\%} \\ \hline
\textbf{\begin{tabular}[c]{@{}l@{}}False \\ negatives\end{tabular}} & \begin{tabular}[c]{@{}l@{}}Is there a mention of \\ fluid in the record?\end{tabular} & \begin{tabular}[c]{@{}l@{}}There is some fluid, or \\ mucosal thickening in \\ the ethmoid and \\ sphenoid sinuses\end{tabular} & \begin{tabular}[c]{@{}l@{}}The amount of fluid \\ layering at the apices \\ and the pleural spaces \\ appear slightly decreased\end{tabular} & \textbf{2\%} & \textbf{18\%} \\ \hline
\textbf{\begin{tabular}[c]{@{}l@{}}May need\\ external \\ knowledge \\\end{tabular}} & \begin{tabular}[c]{@{}l@{}}
What treatment has \\ the patient had \\for his CAD?
\end{tabular} & \begin{tabular}[c]{@{}l@{}}CAD s/p \\ CABG 2003 s/p\end{tabular} & \begin{tabular}[c]{@{}l@{}}Pt's vancomycin was \\stopped after 14\\ days of treatment\end{tabular} & \textbf{2\%} & \textbf{2\%} \\ \hline
\textbf{Others} & \begin{tabular}[c]{@{}l@{}}Is the patient's right \\ hand ganglion cyst \\ well-controlled?\end{tabular} & \begin{tabular}[c]{@{}l@{}}right hand ganglion \\ cyst removed\end{tabular} & \begin{tabular}[c]{@{}l@{}}x 3 right hand \\ ganglion cyst\end{tabular} & \textbf{4\%} & \textbf{8\%} \\ \hline
\end{tabular}%
}
\caption{Error analysis on 50 sampled questions from the \textit{Medication} and \textit{Relation} dev sets respectively. Example question, ground truth and prediction from either \textit{Medication} or \textit{Relation} are given for each type of error.}
\label{tbl:error_analysis}
\vspace{-10pt}
\end{table*}

Since the questions were generated from expert created templates, most of them are human-readable and unambiguous. We therefore mainly focus on evaluating answer quality. We ask two human experts to score each answer from 1 to 5 depending on the relevance of the answer to the question (1: irrelevant or incorrect; 2: missing key parts; 3: contains key parts but is not human-readable or contains many irrelevant parts; 4: contains key parts and is only missing a few parts or has a few irrelevant extra segments; 5: perfect answer). We also ask human annotators to label the gold answers and then calculate the Exact Match (EM) and F1 score (F1) of the emrQA answers v.s. human gold answers. The answer quality score, EM and F1 in both datasets, are shown in Table \ref{tbl:dataset_quality}.

The scores of the \textit{Medication} dataset are low since most of the answers are broken sentences or contain unnecessary segments. For instance, in the Figure \ref{fig:emrQA_generation} example, the correct answer should be \textit{``Clindamycin was changed to Flagyl"}, however, the emrQA answer misses important parts \textit{``Clindamycin was changed to"} and contains irrelevant parts \textit{``By discharge, the patient was afebrile"}. These issues are common in the \textit{Medication} dataset and make it difficult to train a good system. To understand why the generated answers contain such noise, we explored the \textit{``i2b2 2009 Medication"} challenge dataset which was used to create these QA pairs. We found that most documents in this dataset contain many complete sentences split into separate lines. Since the i2b2 annotation are token based and the emrQA obtains full lines around the token as evidence spans, these lines often end up being broken sentences. We tried to relabel the answers with existing sentence segmentation tools and heuristic measures but found that it is very challenging to obtain concise and complete text spans as answers.

Compared with the \textit{Medication} dataset, the answer quality of the \textit{Relation} dataset is much better. In most cases, the answers are complete and meaningful sentences with no unnecessary parts.

\subsection{How challenging are the emrQA pairs?}
\label{subsec:question_difficulty}
Another observation from the 50 samples is that 96\% of the answers in the \textit{Medication} dataset and 100\% of the answers in the \textit{Relation} dataset contain the key phrase in the question. This is due to the generation procedure illustrated in Figure \ref{fig:emrQA_generation}. In this example, the key phrase or entity (\textit{``Flagyl"}) in the question  is also included in the answer. This undoubtedly makes the answer easier to extract as long as the model can recognize significant words and do ``word matching".

To further explore how much clinical language understanding is needed and what kind of errors do the state-of-the-art reader make, we conduct error analysis using DocReader \cite{chen2017reading} (also used in \cite{pampari2018emrqa}) on the emrQA dataset. More specifically, we randomly sample 50 questions that are answered incorrectly by the model (based on exact match metric) from the \textit{Medication} and \textit{Relation} dev set respectively\footnote{Note that these 100 samples are sampled from errors, which are different from the previously sampled ones.}. The results are shown in Table \ref{tbl:error_analysis} (examples for each error type are also given for better understanding). 

Since emrQA answers are often incomplete in the dataset, we deem \textit{span mismatch} errors acceptable as long as the predictions include the key part of the ground truths. Surprisingly, \textit{span mismatch-include key info} errors, along with \textit{ambiguous questions}, \textit{incorrect golds} and \textit{false negatives} (the prediction is correct but it is not in the emrQA answers) errors, which are caused by the dataset itself, account for 90\% of total errors, suggesting that the accuracy of these models is even higher than we report. 

Another interesting finding from the error analysis is that to our surprise, only a very small amount (2\%) of errors may have been caused by a lack of external domain knowledge while \citet{pampari2018emrqa} claim that 39\% of the questions in the emrQA dataset need domain knowledge. This surprising result might be due to: (1) neural models being able to encode relational or associative knowledge from the text corpora as has also been reported in recent studies \cite{petroni2019language,bouraoui2019inducing}, and (2) questions and answers sharing key phrases (as we mentioned earlier in \cref{subsec:answer_quality}) in many samples, making it more likely that fewer questions need external knowledge to be answered than previously reported.

\section{Comprehensive Quantitative Analysis}
In this section, we conduct comprehensive experiments on the emrQA dataset with state-of-the-art readers and recently dominating BERT models. Full experimental settings are described in Appendix \ref{sec:exp_setting}.

\begin{table*}[!t]
\resizebox{\textwidth}{!}{%
\begin{tabular}{lcccccccc}
\hline
\multirow{3}{*}{\textbf{Model}} & \multicolumn{4}{c}{\textbf{Medication}} & \multicolumn{4}{c}{\textbf{Relation}} \\ \cline{2-9} 
 & \multicolumn{2}{c}{Dev} & \multicolumn{2}{c}{Test} & \multicolumn{2}{c}{Dev} & \multicolumn{2}{c}{Test} \\
 & EM & F1 & EM & F1 & EM & F1 & EM & F1 \\ \hline
BiDAF \cite{Seo2017Bidirectional} & 25.50  & 68.13  & 23.35  &67.18  & 81.51 &90.84  &82.74  &91.27  \\
DocReader \cite{chen2017reading} & 29.20 & 72.78 & 25.68 & 70.45 & 86.43 & 94.44 & 86.94 & 94.85 \\
QANet \cite{yu2018qanet} & 27.67 & 69.40 & 24.74 & 67.34 & 82.41 & 90.61 & 82.68  &91.56  \\ \hline
BERT-base \cite{devlin2019bert} & 26.62 & 68.75 & 24.00  &  67.49 & 80.17 & 90.01& 83.29 & 92.38  \\
BioBERT \cite{Lee2019biobert} & 27.81 & 71.90 & 24.75 & 69.97  & 81.57 & 91.38 & 83.61  & 92.62 \\
ClinicalBERT \cite{alsentzer2019publicly} & 27.14 & 71.84 & 24.06 & 69.05 & 83.12 & 91.96 &85.33  &93.06  \\ \hline
\end{tabular}%
}
\caption{Overall performance of all models on the \textit{Medication} and \textit{Relation} dataset. All numbers are percentages.}
\label{tbl:main_results}
\end{table*}

\subsection{How redundant are the emrQA pairs?}
\label{subsec:sample_subset}
Though there are more than 1 million questions in the emrQA dataset (as shown in Table \ref{tbl: statistics}), many questions and their patterns are very similar since they are generated from the same question templates. This observation leads to a natural question: \textit{do we really need so many questions to train an CliniRC system?} If many questions are similar to each other, it is very likely that using a sampled subset can achieve roughly the same performance that is based on the entire dataset.

To verify our hypothesis, we first split the two datasets into train, dev, and test set with the proportion of 7:1:2 w.r.t. the contexts (full statistics are shown in Appendix Table A1). Then we randomly sample \{5\%, 10\%, 20\%, 40\%, 60\%\} and \{1\%, 3\%, 5\%, 10\%, 15\%\}\footnote{The sampling percentage of the \textit{Relation} dataset is smaller than the \textit{Medication} dataset since the former one has more QA pairs (roughly 4 times).} of the QA pairs in each document (context) of the \textit{Medication} and the \textit{Relation} training sets respectively. We run DocReader \cite{chen2017reading}  on the sampled subsets and evaluate them on the same dev and test set.

\begin{figure}
    \centering
    \includegraphics[width=\linewidth]{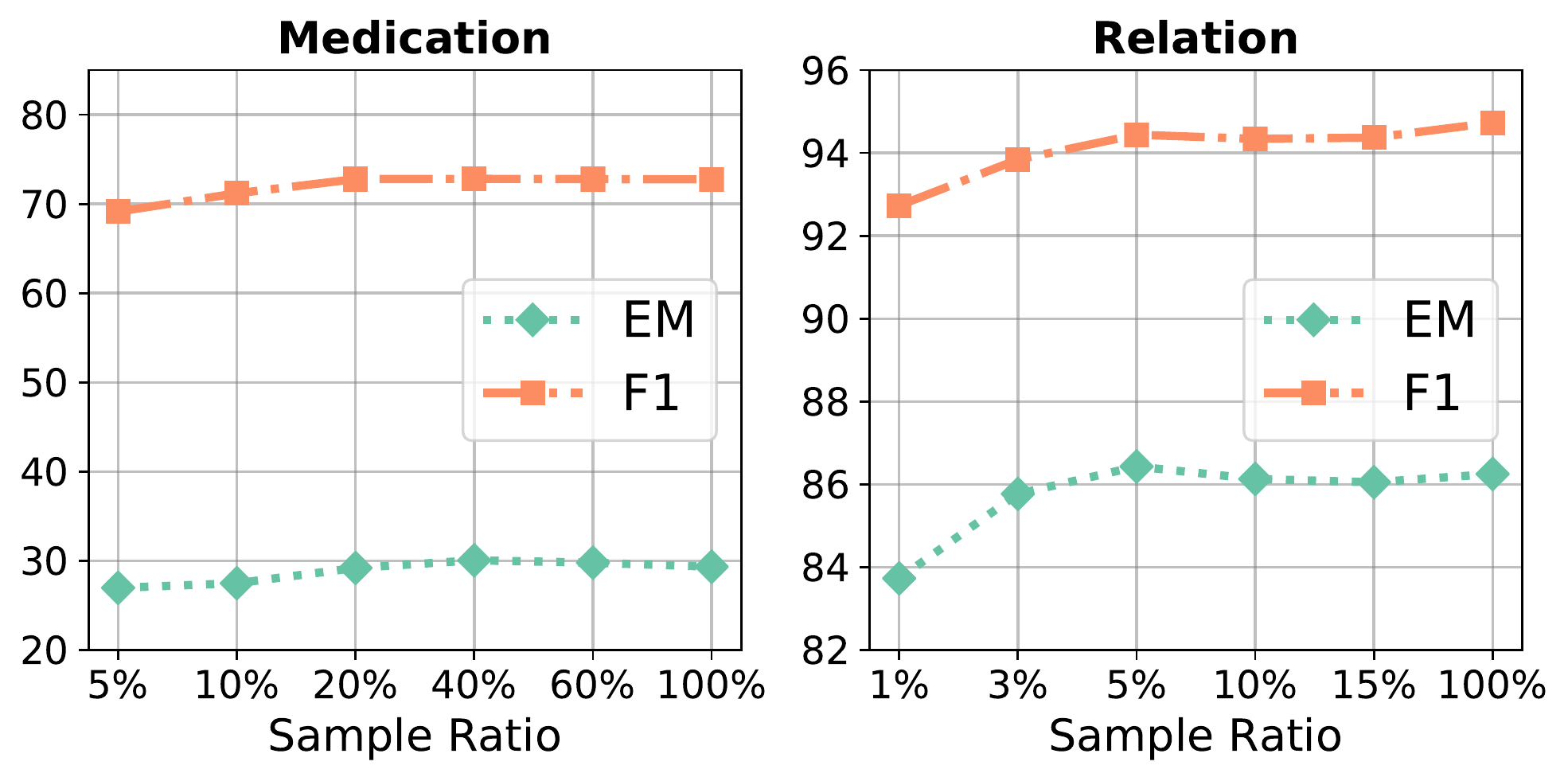}
    \caption{Impact of \textit{training size} on the performance of DocReader \cite{chen2017reading} based on the \textit{Medication} and \textit{Relation} dataset.}
    \label{fig:sampling}
\end{figure}

As shown in Figure \ref{fig:sampling}, using 20\% of the questions in the \textit{Medication} and 5\% of the questions in the \textit{Relation} dataset can achieve roughly the same performance as using the entire training sets. These verify our hypothesis, and illustrate learning a good and robust reader system based on the emrQA dataset does not need so many question-answer pairs. While deep models are often data-hungry, it does not mean more data can always lead to better performance. In addition to the training size, diversity should also be considered as another important criterion for data quality.

In the following experiments, we use the sampled subsets (20\% for \textit{Medication} and 5\% for \textit{Relation}) considering the time and memory cost as well as performance.

\subsection{Little room for improvement}
\label{subsec:ceiling}

Since the answers in emrQA are often incomplete, the performance of a model is more appropriately reflected by its F1 score. As shown in Table \ref{tbl:dataset_quality}, we obtain F1 scores of 74\% and 95\% on two datasets respectively when we test human-labeled answers against the emrQA answers on a sampled dataset. We can see from Table \ref{tbl:main_results} that the best performing reader, DocReader, achieves around 70\% and 94\% F1 performance on the \textit{Medication} and \textit{Relation} test set respectively, {which} are very close to the human performance just described. Though designing more complex and advanced models may achieve better scores, such scores are obtained w.r.t. noisy emrQA answers and may not translate meaningfully to real cases.

\subsection{BERT does not always win}
\label{subsec:worse_BERT}
BERT models have achieved very promising results recently in various NLP tasks including RC \cite{devlin2019bert}. We follow their experiment setting of BERT for doing reading comprehension on the SQuAD \cite{rajpurkar2016squad} dataset. To our surprise, as shown in Table \ref{tbl:main_results}, BERT models (BERT-base, its biomedical version BioBERT \cite{Lee2019biobert}, and its clinical version ClinicalBERT \cite{alsentzer2019publicly}) do not dominate as they do in the open-domain RC tasks. The reasons may be three-fold: 1) BERT benefits the most from large training corpora. The training corpora of BERT-base and BioBERT are Wikipedia + BookCorpus \cite{zhu2015aligning} and PubMed articles respectively, both of which may have different vocabularies and use different language expressions from clinical texts. Though ClinicalBERT was pretrained on MIMIC-III \cite{mimiciii} clinical texts, the training size of the corpus ($\sim$50M words) is far less than that used in BERT ($\sim$3300M words), which may make the model {less} powerful as it is on the open-domain tasks.  2) Longer Contexts. As can be seen from Table \ref{tbl: statistics}, the number of tokens in the contexts is commonly larger than open-domain RC datasets like SQuAD ($\sim$1000 v.s.$\sim$116 avg). We suspect that long contexts might make it more challenging to model sequential information. For sequences that are longer than the \textit{max length} of the BERT model, they are truncated into a set of short sequences, which may hinder the model from capturing long dependencies \cite{dai-etal-2019-transformer} and global information in the entire document. 3) Easy Questions. Another possible reason might be the question patterns are too easy and a simpler reader with far less parameters can learn the patterns and obtain satisfying performance.

Additionally, to further evaluate the models in the fine-grained level, inspired by \cite{gururangan2018annotation}, we partition the \textit{Medication} and \textit{Relation} test sets into Easy and Hard subsets using a base model. The details of Easy/Hard splits can be found in Appendix \ref{sec:question_split}. As can be seen from Table \ref{tbl:easy_hard}, most of the questions in the two datasets are easy, which indicates the emrQA dataset might not be challenging for the current QA models. More difficult datasets are needed to advance the Clinical Reading Comprehension task.

\section{Desiderata in Real-World CliniRC}
Following our analysis of the emrQA dataset, we further study two aspects of clinical reading comprehension systems that we believe are crucial for their real-world applicability: the need to encode clinical domain knowledge and to generalize to unseen questions and documents.

\subsection{External domain knowledge is needed}
\label{subsec:external_knowledge}
So far, we have shown that domain knowledge may not be very useful for models answering questions \textit{in the emrQA dataset}; however, we argue that systems in real-world CliniRC need to be able to encode and use clinical domain knowledge effectively.

Clinical text often contains high variability in many domain-specific words due to abbreviations and synonyms. The presence of different aliases in the question and context can make it difficult for a model to represent semantics accurately and choose the correct span. Besides, medical domain-specific relations (e.g., \textit{treats, caused by}) and hierarchical relations (e.g., \textit{isa}) between medical concepts would be likely to appear. The process followed to generate the current emrQA dataset leads to these problems being largely under-represented, even though they can be very common in real cases. We use the following 3 examples as representatives to illustrate the real cases we may encounter.

\noindent \textbf{Synonym.} For example, for the question in Figure \ref{fig:emrQA_generation}, \textit{``Has this patient ever been on Flagyl?"}, it is easy for the model to answer since \textit{``Flagyl"} appears in the context. However, if we change \textit{``Flagyl"} to its synonyms \textit{``Metronidazole"} (which may not appear in training) in the question, it is hard for the reader to extract the correct answer, as it is not possible for model to capture the semantic meaning of \textit{``Metronidazole"}  as \textit{``Flagyl"}. 

\noindent \textbf{Clinical Relations.} Another example is the question shown in Figure \ref{fig:intro_example}, \textit{``Why has the patient been prescribed hctz?"}. Currently, machines can easily find the answer since keyword \textit{``hctz"} is mentioned in the answer. However, given a situation where the drug \textit{``hctz"} does not appear in the local context of \textit{``HTN"}, our model may have a better chance to extract the correct answers if it stores the relation \textit{``(hctz, treats, HTN)"}.

\noindent \textbf{Hierarchical Relation.} For the question \textit{``Is there a history of mental illness?"}, it is more likely that the medical report describes a specific type of psychological condition rather than mention the general phrase \textit{``mental illness"} since clinical support require specifics. To obtain the correct answer in this case \textit{``Depression with previous suicidal ideation."}, encoding the relation \textit{``(depression, isa, mental illness)"} would probably help the model make a correct prediction.

These three cases help illustrate how complex medical relations affect the real CliniRC task. Without leveraging external domain knowledge, it is difficult for models to capture the semantic relations necessary to resolve such cases.

In order to verify our claim quantitatively, we select \textit{synonym} as a representative relation type and manipulate each question by replacing its entities with plausible synonyms or abbreviations. We then introduce external domain knowledge into current models and compare their performance against base models on these augmented questions. 

More specifically, we first detect entities in the questions and link them to a medical knowledge base (KB): UMLS \cite{bodenreider2004unified} using a biomedical and clinical text NLP pipeline tool, \textit{ScispaCy} \cite{neumann2019scispacy}. Synonyms of detected entities are then retrieved from UMLS and used to replace the original mention. We filter the questions that do not contain entities or that contain entities with no synonyms. We focus on the \textit{Relation} dataset and only modify the questions in the dev and test set; the questions in the training set are not modified. Finally, we get 69,912 and 125,338 questions in the dev and test set.

We then introduce a simple Knowledge Incorporation Module (KIM) to evaluate the usefulness of external domain knowledge. Formally, given a question $q: \{w^q_1, w^q_2, ..., w^q_l\}$ and its context $c: \{w^c_1, w^c_2, ..., w^c_m\}$, where $w^q_i, w^c_j$ are words (tokens), all the words can be mapped to $d_1$ dimensional vectors via a word embedding matrix $E_w \in \mathbb{R}^{d_1 \times |\mathcal{V}|}$, where $\mathcal{V}$ denotes the word vocabulary.
So we have $q: \mathbf{w^q_1,..., w^q_l} \in \mathbb{R}^{d_1}$ and $c: \mathbf{w^c_1,..., w^c_m} \in \mathbb{R}^{d_1}$.

We then detect entities $\{e^q_1, e^q_2,...,e^q_n\}$ in the question and entities $\{e^c_1, e^c_2,...,e^q_o\}$  in the context and map them to a medical knowledge base (KB), UMLS \cite{bodenreider2004unified} using scispacy \cite{neumann2019scispacy}. Note that $l$ is not equal to $n$ and $m$ is not equal to $o$, since not every token can be mapped to a entity in KB. For entities that contain multiple words, we align them to the first token, same as the alignment used in \cite{zhang2019ernie}. We then map detected entities to $d_2$ dimensional vectors  $\mathbf{\{e^q_1, e^q_2,...,e^q_n\}}$ and $\mathbf{\{e^c_1, e^c_2,...,e^c_o\}}$ via a entity embedding matrix $E_e \in \mathbb{R}^{d_2 \times |\mathcal{U}|}$ , which is pretrained on the entire UMLS KB using the knowledge embedding method TransE \cite{bordes2013translating}. $\mathcal{U}$ denotes the entity vocabulary. 

We merge the word embeddings with entity embeddings to feed them into a Multi-layer Perceptron (MLP):
\begin{equation}
\begin{aligned}
        \mathbf{h^q_i} &= \sigma (\mathbf{W_c w^q_i + W_e e^q_i +b})\\
        \mathbf{h^c_j} &= \sigma (\mathbf{W_c w^c_j + W_e e^c_j +b})
\end{aligned}
\end{equation}
where $\sigma$ is activation function, $W_c, W_e, b$ are trainable parameters and $h^q_i, h^c_j$ denote the integrated embeddings that contain information from both the word $c_j$ and the entity $e_j$ in the question and context respectively. For the word that is not mapped to an entity, $e_j$ will be set to $\mathbf{0}$. The merged embeddings are used as the input to the base reader. 

\begin{figure}[t]
    \centering
    \includegraphics[width=\linewidth]{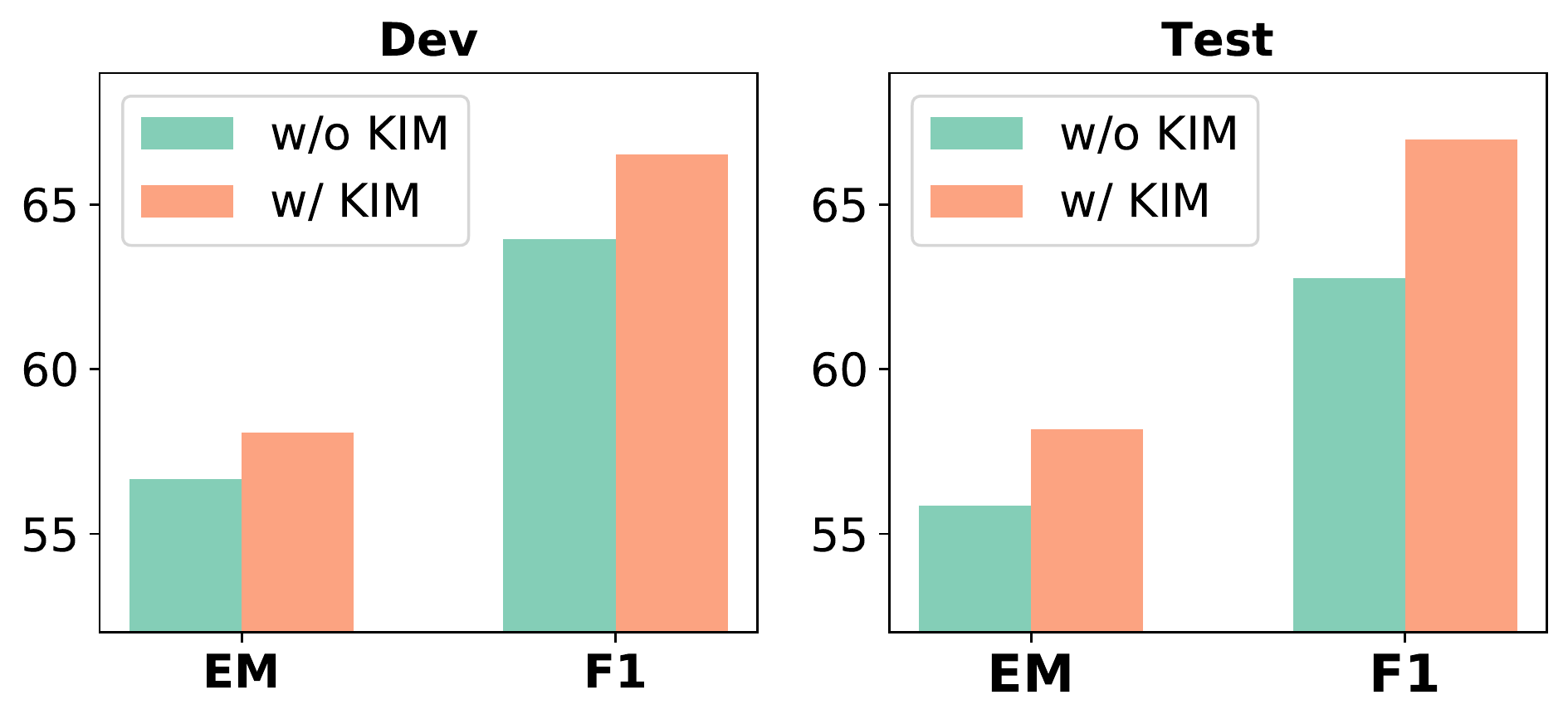}
    \caption{Performances of DocReader and DocReader + Knowledge Incorporation Module (KIM) on our created questions modified from the \textit{Relation} dataset.}
    \vspace{-10pt}
    \label{fig:knowledge}
\end{figure}

As shown in Figure \ref{fig:knowledge},  by adding a basic Knowledge Incorporation Module to the base model, we obtain around 5\% increase of F1 score on the manipulated questions in the test set. This suggests that for questions that involve relations between medical concepts, external domain knowledge may be quite important. 

\begin{table*}[t]
\centering
\begin{tabular}{ccccccccc|cc}
\hline
Model        & \multicolumn{2}{c}{\begin{tabular}[c]{@{}c@{}}Existing \\ Questions \end{tabular}} & \multicolumn{2}{c}{\begin{tabular}[c]{@{}c@{}}Paraphrased\\ Questions\end{tabular}} & \multicolumn{2}{c}{\begin{tabular}[c]{@{}c@{}}New \\ Questions\end{tabular}} & \multicolumn{2}{c}{\begin{tabular}[c]{@{}c@{}}Overall\end{tabular}}&
\multicolumn{2}{|c}{\begin{tabular}[c]{@{}c@{}}emrQA\\ Relation\end{tabular}}  \\ \hline
             & EM       & F1       & EM 
             & F1        & EM     & F1 
             & EM     & F1      & EM     & F1                         \\ \cline{2-11} 
DocReader   & 58.33    & 71.62    & 38.09      & 57.28     & 29.41   & 35.35 
& 40.00
& 53.27
& 86.94
& 94.85
\\
ClinicalBERT & 58.33    & 73.12    & 38.09      & 62.04    & 23.53  & 48.79 
& 38.00
& 60.19
& 85.33
& 93.06
\\   \hline                  
\end{tabular}

\caption{Results of models when tested on new questions and unseen clinical notes (not in emrQA, but from MIMIC-III dataset). Performance drops around 40\% compared with previously reported on the \textit{Relation} test set, highlighting generalizability as an essential future direction for CliniRC.}
\vspace{-10pt}
\label{tbl:mimiciii}
\end{table*}

\subsection{Generalizing to unseen questions and documents}
\label{subsec:generalizability}
The aim of CliniRC is to build robust QA systems for doctors to retrieve information buried in clinical texts. When deploying a CliniRC system to a new environment (e.g., a new set of clinical records, a new hospital, etc.), it is infeasible to create new QA pairs for training every time. Thus, an ideal CliniRC system is able to generalize to unseen documents and questions after being fully trained.

To test the generalizability of models trained on emrQA (we focus on the \textit{Relation} dataset here), our medical experts created 50 new questions that were not present in the emrQA dataset and extracted answers from unseen patient notes in the MIMIC-III \cite{mimiciii} dataset. This dataset consists of three types of questions: 12 questions were made from emrQA question templates but contain entities which do not appear in the training set (e.g., \textit{``How was the diagnosis of acute cholecystitis made?"} was created from the template \textit{``How was the diagnosis of $|$problem$|$ made?"}). The other 38 questions have different forms from existing question templates: 21 paraphrase existing questions from emrQA (e.g., \textit{``Was an edema found in the physical exam?")} was paraphrased from \textit{``Does he have any evidence of $|$problem$|$ in $|$test$|$?"}) and 17 are completely semantically different from the ones in the emrQA dataset (e.g., \textit{``What chemotherapy drugs are being administered to the patient?"}).

As could be expected, we see in Table \ref{tbl:mimiciii} that the more the new questions deviate from the original emrQA, the more the models struggle to answer them. We observe a performance drop of roughly 20\% compared to the \textit{Relation} test set on questions made from emrQA templates using MIMIC III clinical notes which were not in the original dataset. For question that are more significantly different, we notice an approximate 40\% and 60\% loss in F1 score when predicting paraphrased questions and entirely new questions respectively. This steep drop in performance for these new settings, especially for paraphrased and new questions, shows how much work there is to be done on this front and highlights generalizability as an important future direction in CliniRC. We also notice that ClinicalBERT works slightly better than the base model DocReader. The reason might be ClinicalBERT was pretrained on the MIMIC-III dataset, which might help the model have a better understanding of the context. 

~\\
\noindent \textbf{Summary.} Based on these two aspects and our previous thorough analysis of the emrQA dataset, it is clear that better datasets are needed to advance CliniRC. Such datasets should be not only large-scale, but also less noisy, more diverse, and moreover allow researchers to systematically evaluate a model's ability to encode domain knowledge and to generalize to new questions and contexts.

\section{Related Work}
We present a brief overview of open-domain, biomedical and clinical question answering tasks, which are most related to our work: 

\textbf{Question Answering (QA)} aims to automatically answer questions asked by humans based on external sources, such as Web \cite{sun2016table}, knowledge base \cite{yih2015semantic,sun2015open} and free text \cite{chen2016thorough}. As an important type of QA, reading comprehension intends to answer a question after reading the passage \cite{hirschman1999deep}. Recently, the release of large-scale RC datasets, such as CNN \& Daily Mail \cite{hermann2015teaching}, Stanford Question-Answering Dataset (SQuAD) \cite{rajpurkar2016squad,rajpurkar2018know} makes it possible to solve RC tasks by building deep neural models \cite{hermann2015teaching, Wang2016MachineCU, Seo2017Bidirectional, chen2017reading}.

More recently, contextualized word representations and pretrained language models, such as ELMo \cite{peters2018deep}, GPT \cite{radford2018improving}, BERT \cite{devlin2019bert}, have been demonstrated to be very useful in various NLP tasks including RC. By seeing diverse contexts in large corpora, these pretrained language models can capture the rich semantic meaning and produce more accurate and precise representations for words given different contexts. Even a simple classifier or score function built upon these pretrained contextualized word representations perform well in extracting answer spans \cite{devlin2019bert}. 

\noindent \textbf{Biomedical and Clinical QA}. Due to the lack of large-scale annotated biomedical or clinical data, QA and RC systems in these domains are often rule-based and heuristic feature-based \cite{lee2006beyond,niu2006using,athenikos2010biomedical}. 

In recent years, BioASQ challenges \cite{Balikas2015bioasq} proposed the Biomedical Semantic QA task, where the participants need to respond to each test question with relevant articles, snippets and exact answers. \citet{vsuster2018clicr} use summary points of clinical case reports to build a large-scale cloze-style dataset (CliCR), which is similar to the style of CNN \& Daily Mail dataset. \citet{jin2019pubmedqa} presents PubMedQA, which extracts question-style titles and their corresponding abstracts as the questions and contexts respectively. A few QA pairs are annotated by human experts and most of them are annotated based a simple heuristic rule with ``yes/no/maybe". 

Due to the great power of contextualized word representations,  pretrained language models also have been introduced to biomedical and clinical domain, e.g., BioELMo \cite{jin2019probing}, BioBERT \cite{Lee2019biobert}, and ClinicalBERT \cite{alsentzer2019publicly}. They adopt similar architectures of the original models but pretrained on the medical and clinical corpus, such as PubMed articles and MIMIC-III \cite{mimiciii} clinical notes. 

\section{Conclusion}
We study the Clinical Reading Comprehension (CliniRC) task with the recently created emrQA dataset. Our qualitative and quantitative analysis as well as exploration of the two desired aspects of CliniRC systems show that future clinical QA datasets should not only be large-scale but also less noisy and more diverse. Moreover, questions that involve complex relations and are across different domains should be included, and then more advanced external knowledge incorporation methods as well as domain adaptation methods can be carefully designed and systematically evaluated.

\section*{Acknowledgments}
We thank our medical experts for their annotations. We thank Ping Zhang, Changchang Yin and anonymous reviewers for their helpful comments. This research was sponsored in part by the Patient-Centered Outcomes Research Institute Funding ME-2017C1-6413, the Army Research Office under cooperative agreements W911NF-17-1-0412, NSF Grant IIS1815674, and Ohio Supercomputer Center \cite{OhioSupercomputerCenter1987}. The views and conclusions contained herein are those of the authors and should not be interpreted as representing the official policies,
either expressed or implied, of the Army Research Office or the U.S.Government. The U.S. Government is authorized to reproduce and distribute reprints for Government purposes notwithstanding any copyright notice herein.


\clearpage
\appendix
\section{Experimental Set-up}
\label{sec:exp_setting}

\setcounter{table}{0}
\setcounter{footnote}{0}
\renewcommand{\thetable}{A\arabic{table}}
We split the two datasets \textit{Medication} and \textit{Relation} based on the documents (clinical texts) into train, dev, test with the ratio 7:1:2. The statistics are shown in Table \ref{tbl:stat_train_dev_test}.

We adopt Exact Match (EM) and F1 score (F1) as our evaluation metrics, same as the open-domain RC \cite{rajpurkar2016squad}. We use SQuAD v1.1 official evaluation script \footnote{https://rajpurkar.github.io/SQuAD-explorer/} to evaluate all the models.   All the models used in the paper, BiDAF \footnote{https://github.com/allenai/bi-att-flow}, DocReader \footnote{https://github.com/facebookresearch/DrQA}, QANet \footnote{https://github.com/BangLiu/QANet-PyTorch}, BERT \footnote{https://github.com/google-research/bert}, BioBERT \footnote{https://github.com/dmis-lab/biobert}, ClinicalBERT \footnote{https://github.com/EmilyAlsentzer/clinicalBERT} are run based on the implementations listed here and strictly followed the instructions.

For reproducibility, we list all the key hyperparaters we use for each method in the Table \ref{tbl:hyperparameters}.

We implement our Knowledge Incorporation Module based on DocReader implementations. Entity embeddings are pretrained using TransE \cite{bordes2013translating} with the dimension of 100.  The hyperparameters are kept same as the DocReader. All the models are run on NVIDIA GeForce GTX 1080 GPUs. We save the best model (with the highest EM) on the dev set and use it for test set.

\begin{table}[t]
\begin{tabular}{lcc}
\hline \hline
 & \textbf{Medication} & \textbf{Relation} \\ \hline
\textbf{\# Train (Q / C)} & 154,684 /182 & 621,428 / 296 \\
\textbf{\# Dev (Q / C)} & 23,081 / 26 & 101,700 / 42 \\
\textbf{\# Test (Q / C)} & 45,192 / 53 & 181,464 / 85 \\
\textbf{Total} & 222,957 / 261 & 904,592 / 423 \\ \hline
\end{tabular}
\caption{Statistics of train, dev, test set of the \textit{Medication} and \textit{Relation} datasets.}
\label{tbl:stat_train_dev_test}
\end{table}

\begin{table}[t]
\resizebox{\linewidth}{!}{%
\begin{tabular}{ll}
\hline
\textbf{Method} & \textbf{Hyper-parameters Setting} \\ \hline \hline
\textbf{DocReader} & \begin{tabular}[c]{@{}l@{}}epoch: 30; batch-size: 16; \\ test-batch-size:16; droput-rate:0.4; \\ doc-layers: 3; question-layers: 3; \\  grad-clipping: 10; \\ tune-partial: 1000; max-len: 30; \\ the others are set as default\end{tabular} \\ \hline
\textbf{BiDAF} & \begin{tabular}[c]{@{}l@{}}init\_lr: 0.001; batch-size:6; \\ num\_epochs: 2; cluster: True; \\ len-opt: True; word\_count\_th: 10; \\ char\_count\_th: 50; sent\_size\_th: 4000; \\ num\_sents\_th: 500; ques\_size\_th: 30; \\ word\_size\_th: 30; para\_size\_th: 4000; \\ the others are set as default;\end{tabular} \\ \hline
\textbf{QANet} & \begin{tabular}[c]{@{}l@{}}batch-size: 4; lr: 0.001; \\ grad-clip: 5; use-ema: True;\\ epoch: 30; para\_limit: 4000; \\ ques\_limit: 30; ans\_limit: 30; \\ char\_limit: 40; num-head:1;\\ the others are set as defalut;\end{tabular} \\ \hline
\textbf{\begin{tabular}[c]{@{}l@{}}BERT-base\\ BioBERT\\ ClinicalBERT\end{tabular}} & \begin{tabular}[c]{@{}l@{}}train\_batch\_size: 6; \\ learning\_rate: 3e-5; \\ num\_train\_epochs: 3.0;\\ max\_seq\_length: 384; \\ doc\_stride: 128;\\ the others are set as default;\end{tabular} \\ \hline
\end{tabular}%
}
\caption{Hyperparameters settings for all the  methods used in the experiments.}
\label{tbl:hyperparameters}
\end{table}

\begin{table}[t]
\resizebox{\linewidth}{!}{%
\begin{tabular}{cccccc}
\hline
\multirow{2}{*}{Dataset} & \multirow{2}{*}{Model} & \multicolumn{2}{c}{Dev} & \multicolumn{2}{c}{Test} \\ \cline{3-6} 
 &  & EM & F1 & EM & F1 \\ \hline
\multirow{2}{*}{medication} & DocReader & 32.19 & 76.21 & 33.45 & 77.08 \\
 & ClinicalBERT & 30.16 & 74.81 & 32.18 & 75.79 \\ \hline
\multirow{2}{*}{relation} & DocReader & 87.21 & 94.32 & 87.54 & 94.97 \\
 & ClinicalBERT & 85.46 & 93.92 & 85.67 & 93.14 \\ \hline
\end{tabular}%
}
\caption{Performance of the two models on the shorter context setting.}
\label{tbl:short_context}
\end{table}

\begin{table*}[t]
\resizebox{\linewidth}{!}{%
\begin{tabular}{c|c|ccccccc}
\hline
\multicolumn{2}{c|}{\multirow{3}{*}{\begin{tabular}[c]{@{}c@{}}\textbf{Distribution of} \\ \textbf{Easy/Hard Questions}\end{tabular}}} &  & \multicolumn{2}{c}{Easy} & \multicolumn{2}{c}{Hard} & \multicolumn{2}{c}{Total} \\ \cline{3-9} 
\multicolumn{2}{c|}{} & Medication & \multicolumn{2}{c}{33,037 (73.1\%)} & \multicolumn{2}{c}{12,155 (26.9\%)} & \multicolumn{2}{c}{45,192 (100\%)} \\
\multicolumn{2}{c|}{} & Relation & \multicolumn{2}{c}{165,271 (91.1\%)} & \multicolumn{2}{c}{16,193 (8.9\%)} & \multicolumn{2}{c}{181,464 (100\%)} \\ \hline\hline
\multirow{6}{*}{\textbf{Results}} & \multirow{2}{*}{} & \multirow{2}{*}{\textbf{Model}} & \multicolumn{2}{c}{Easy} & \multicolumn{2}{c}{Hard} & \multicolumn{2}{c}{Total} \\ \cline{4-9} 
 &  &  & EM & F1 & EM & F1 & EM & F1 \\ \cline{2-9} 
 & \multirow{2}{*}{\textbf{Medication}} & DocReader \cite{chen2017reading}   & 30.25 & 73.78 & 13.26 & 61.46 & 25.68 & 70.45 \\
 &  & ClinicalBERT \cite{alsentzer2019publicly} & 28.25 & 72.02 & 12.64 & 60.98 & 24.06 & 69.05 \\ \cline{2-9} 
 & \multirow{2}{*}{\textbf{Relation}} & DocReader \cite{chen2017reading}   & 87.66 & 95.39 & 79.85 & 89.62 & 86.94 & 94.85 \\
 &  & ClinicalBERT \cite{alsentzer2019publicly} & 86.06 & 93.71 & 78.09 & 86.57 & 85.33 & 93.06 \\ \hline
\end{tabular}%
}
\caption{Performance of DocReader and ClinicalBERT on the easy/hard questions split.}
\label{tbl:easy_hard}
\end{table*}

\section{Performance on Shorter Contexts}
\label{sec:short_context}

Using the entire clinical record as the context might be too long for models to capture sequential information. We also try to split the entire record into different sections (e.g., ``medical history", ``family history") based on some heuristic measures. Specifically, in order to split the clinical notes into sections, we notice that most sections begin with easily identifiable headers. To detect these headers we use a combination of heuristics such as whether the line contains colons, all uppercase formatting or phrases found in a list of clinical headers taken from SecTag \cite{denny2009evaluation}. We then select the section that contains the answer as the context ($\sim$100 words avg). We select DocReader and ClinicalBERT as representative methods and re-run them on the modified shorter context. The results are shown in Table \ref{tbl:short_context}. The performance of the two models is improved compared with the performance of models built on the whole record (long context). However, ClinicalBERT still does not outperform DocReader in this setting, indicating that longer context may not explain why BERT models do not win on this dataset or that shortening context in a such manner might break long dependencies.

This experiment setting may also inspire future research on ``Open Clinical Reading Comprehension". Given that patients often have multiple clinical records, it may not be feasible to jointly use all of them as context for one question. Given multiple records for one patient (instead of just one) and a question, the model would first need to retrieve the most relevant paragraphs and do reading comprehension on each of them or find clever ways to merge them. Such a setting would be interesting for future CliniRC datasets to explore.

\section{Easy/Hard Questions Split}
\label{sec:question_split}
We partition the questions into Easy and Hard. Specifically, we first train a BiLSTM reader and do the prediction on the test set. We obtain the performance of each question template by averaging the performance of all the questions made by this template (such template and question mappings are included in the emrQA dataset). Question templates that obtain higher performance than the overall performance are labeled as "Easy" and "Hard" otherwise. Then we map the difficulty level of question templates back to each question. The reason why we focus on splitting on the question template level is that we can avoid some random noise (e.g., random errors produced by the model on some questions). Also, we release the difficulty level of each question template so that users can easily know which questions are easy or hard and do not need to run a base model to obtain such mappings again. Distributions of easy/hard questions and results of the two selected models are shown in Table \ref{tbl:easy_hard}.

\end{document}